\def\year{2019}\relax
\documentclass[letterpaper]{article} 

\usepackage{aaai19}  
\usepackage{times}  
\usepackage{helvet}  
\usepackage{courier}  
\usepackage{url}  
\usepackage{graphicx}  
\usepackage{amsfonts}
\usepackage{amsmath}
\begin{document}
%

\title{Probabilistic Model Checking of Robots Deployed in Extreme Environments}

\author{Xingyu Zhao\textsuperscript{1}, Valentin Robu\textsuperscript{1, 2},  David Flynn\textsuperscript{1}, Fateme Dinmohammadi\textsuperscript{1}\\ \bf \Large Michael Fisher\textsuperscript{3}, Matt Webster\textsuperscript{3} \\ \textsuperscript{1}School of Engineering \& Physical Sciences, Heriot-Watt University, Edinburgh EH14 4AS, U.K.\\ \textsuperscript{2}Center for Collective Intelligence, Massachusetts Institute of Technology, Cambridge MA 02139, U.S.
\\ \textsuperscript{3}Department of Computer Science, University of Liverpool, Liverpool L69 3BX, U.K.\\
Email:\{xingyu.zhao,\ v.robu,\ d.flynn,\ f.dinmohammadi\}@hw.ac.uk, \{mfisher, matt\}@liverpool.ac.uk}

\maketitle
\begin{abstract}
Robots are increasingly used to carry out critical missions in extreme environments that are hazardous for humans. This requires a high degree of operational autonomy under uncertain conditions, and poses new challenges for assuring the robot's safety and reliability. In this paper, we develop a framework for probabilistic model checking on a layered Markov model to verify the safety and reliability requirements of such robots, both at pre-mission stage and during runtime. Two novel estimators based on conservative Bayesian inference and imprecise probability model with sets of priors are introduced to learn the unknown transition parameters from operational data. We demonstrate our approach using data from a real-world deployment of unmanned underwater vehicles in extreme environments.
\end{abstract}

\section{Introduction}
\label{sec_introduction}
Extreme environments, as a term used by UK EPSRC\footnote{https://epsrc.ukri.org/files/funding/calls/2017/raihubs/} to denote environments that are remote and hazardous for humans, are one of the most promising application areas in which robots can be deployed to carry out a task, such as inspecting oil and gas equipment, maintaining offshore wind turbines or monitoring nuclear reactors. Interaction with human operators is often infeasible in such remote environments, so autonomy becomes a key requirement, meaning robots have to learn to adapt when performing tasks in changing and unexpected circumstances \cite{lane_new_2016}. 

However, unforeseen behaviour of a robot could result either in failure of its own mission or even undermine the integrity of the asset being inspected or repaired -- with potentially catastrophic consequences, e.g. accidental puncture of subsea pipelines. The robots themselves are high-value assets, whose loss or damage may be very costly. 
Moreover, there are increasing demands on regulating autonomous robots to build public trust in their use. Yet, the analysis of safety and reliability for autonomous robots poses a significant challenge due to the inevitable uncertainties in a mission. Potential sources of risk range from failures of sensors or hardware, to built-in algorithms making poor choices in a stochastic environment.
Key industrial foresight reviews~\cite{lane_new_2016} outline a vision of \emph{self-certifying} robotic systems, i.e. systems that continuously monitor their current and predicted performance and assess it within a predefined certification framework \cite{robu_2018_train,fisher_verifiable_2018}.

Probabilistic model checking (PMC) \cite{kwiatkowska_probabilistic_2018} has been successfully used to analyse quantitative properties of systems across a variety of application domains, including robotic systems \cite{luckcuck2018formal}.
 This involves the construction of a probabilistic model, commonly using Discrete Time Markov Chain (DTMC), Continuous Time Markov Chain (CTMC) and Markov Decision Process (MDP) when considering non-deterministic actions, that formally represents the behaviour of a system over time.  
 The properties of interest are normally specified in Linear Temporal Logic (LTL) or Probabilistic Computational Tree Logic (PCTL), then systematic exploration and analysis is performed to check if a claimed property holds.

One inherent problem for most (if not all) formal verification techniques is that the verification assumes the formal model (e.g. DTMC) accurately reflects the actual behaviour of the real-world system \cite{calinescu_2016_formal}. Comparing to conventional systems - for which we might argue the formal model is fairly accurate -- it becomes a tougher issue for systems in changing, unexpected environments and with autonomous features. To handle the issue, the runtime verification idea was proposed \cite{epifani_model_2009,calinescu_2012_self-adaptive} by keeping the formal model alive and continuously updated when seeing new data at runtime. 

In this paper, we propose a tailored PMC framework for verifying safety-critical robots working in extreme environments. Firstly, we formalise how the robot works as a layered and parametric DTMC (transition probabilities are unknown parameters), then feed two Bayesian estimators for different types of parameters with operational data. For each mission, safety and reliability properties are verified at both:

\indent \textbf{\textit{Pre-mission}}, based on the best knowledge from previous similar missions, lab-simulations and experts, which provides assurance before the robot undertaking a mission. 

\indent \textbf{\textit{Runtime}} (i.e. during a running mission), using a DTMC updated in real time. This allows the robot to take protection actions (e.g. restart with new control policies or abort the current mission) whenever a property of interest is being violated, which offers an additional control layer, independent of the robot's front-end planning engine. This follows the ``defence in depth'' design paradigm for safety critical systems, by providing an extra and diverse layer of protection.

To achieve these, we identify two types of transition parameters and introduce two novel Bayesian estimators as:

\emph{A. Catastrophic failure related parameters}, which represent the probability of seeing a catastrophic failure when the robot is in an unsafe state. In practice, we cannot observe sufficient data of catastrophic failures to provide good estimations, since even if we do observe any, we normally will redesign/update the robot making the failure data obsolete. So effectively, for these model parameters, we will only collect \textit{failure-free} data that lead to very optimistic estimations by existing Bayesian estimators. To avoid underestimating the chances of catastrophic failure, our claim is inference in such settings has to be carried out in a \textit{conservative} way. The conservative Bayesian inference (CBI) method \cite{bishop_toward_2011,strigini_software_2013,zhao_conservative_2015,zhao_modeling_2017} was developed for safety-critical software to answer the question what can be claimed rigorously about the reliability when seeing failure-free runs. To our knowledge, our work is the first to develop a CBI estimator for catastrophic failure parameters in robotics.

\emph{B. Non-catastrophic failure related parameters}, which represent the transitions among normal, unsafe and non-catastrophic states, thus sufficient data can be collected as more missions are conducted. Bayesian methods yielding point estimates, e.g. \cite{epifani_model_2009,calinescu_adaptive_2014}, are affected by unquantified estimation errors which will be propagated and compounded in the verification in ways that are unknown but likely to be significant, as highlighted in \cite{calinescu_2016_formal}. Here we introduce an imprecise probability model with \textit{sets of priors} \cite{walter_imprecision_2009,walter_bayesian_2017} to (i) get upper and lower bounds on the posterior estimates whose range measures the estimation errors; (ii) allow modelling imperfect prior knowledge/data from experts/lab-simulations in a flexible way; and (iii) detect \textit{prior-data conflict} \cite{evans_checking_2006} when observe surprising data in a mission to provide protection from the robot's epistemic limits.

We illustrate our new method with an example of an unmanned underwater vehicle (UUV) in the context of a valve turning scenario from the PANDORA\footnote{http://persistentautonomy.com/?p=1436} project which created UUVs that keep going under extreme uncertainty.

Next, we present the necessary background concepts. Two new Bayesian estimators are described in Section \ref{sec_new_Bayesian_estimators}; Section \ref{sec_frame_example} describes the new framework with an illustrative example. Section \ref{sec_related_work} summarises the related work. Contributions, limitations and future work are concluded in Section \ref{sec_conclusion}.

\section{Preliminaries}
\label{sec_prelinimary}

\subsection{DTMC and PCTL}
Discrete Time Markov Chain (DTMC) is a widely-used model for formalising stochastic systems. From the verification point of view, DTMC can serve as a secondary protection model \textit{given} an optimal policy (specifying a procedure for autonomous action selection). For instance, if the primary planner synthesises an optimal policy via MDP and reinforcement learning \cite{PathakShashank2018Varo,pathak_ensuring_2013}, then given the optimal policy, we obtain an \textit{induced DTMC} \cite{puterman_markov_2014} which can be revised to emphasise the safety and reliability aspects by, e.g. adding more transitions to states representing hazards.

\textbf{Definition 1}. A DTMC is a tuple $(S,s_1,\textbf{P},L)$, where:
\begin{itemize}
\item $S$ is a (finite) set of states; and $s_1\in S$ is an initial state;
\item $\textbf{P}:S\times S \rightarrow [0,1]$ is a probabilistic transition matrix such that $\sum_{s^{\prime}\in S}\textbf{P}(s,s^\prime)=1$ for all $s\in S$;
\item $L:S\rightarrow 2^{AP}$ is a labelling function assigning to each state a set of atomic propositions from a set $AP$.
\end{itemize}

We use the notation $p_{ij}=\textbf{P}(s_i,s_j)$, and $i,j$ are integers in $[1,k]$ by assuming there are $k$ states in total. Given an optimal probabilistic\footnote{Deterministic policy is a special case with probability 1.} policy, the transition probability of the induced DTMC is defined as the total probability of:
\begin{equation}
p_{ij}=\sum_{a\in A} \pi_a(s_i)\cdot Pr(s_j \mid s_i,a)
\label{eq_mdp_to_dtmc}
\end{equation}
where $\pi_a(s_i)$ is the probability of executing action $a$ in state $s_i$ according to policy $\pi$ and $Pr(s_j|s_i,a)$ is the probability that $s_j$ is the next state when action $a$ is executed in state $s_i$. In this paper, we assume the optimal policy $\pi$, i.e. $\pi_a(s_i)$, is given by a separate planner and $Pr(s_j|s_i,a)$ will be Bayesian updated via operational data.

The safety and reliability properties to be checked can be specified in Probabilistic Computation Tree Logic (PCTL).

\textbf{Definition 2}. $AP$ is a set of atomic propositions and $ap \in AP, p\in [0,1], t\in \mathbb{N}$ and $\bowtie \in \{<,\leq,>,\geq\}$. The syntax of PCTL is defined by \textit{state formulae} $\Phi$ and \textit{path formulae} $\Psi$.
\begin{align}
\Phi &::= true \mid ap \mid \Phi \wedge \Phi \mid \neg \Phi \mid \mathcal{P}_{\bowtie p}(\Psi) \nonumber
\\
\Psi &::= X\Phi \mid \Phi U^{\leq t} \Phi \nonumber
\end{align}
where the temporal operator $X$ is called \textit{Next} and $U$ is called \textit{Until}. State formulae $\Phi$ is evaluated to be either true or false in each state. Satisfaction relations for a state $s$ are defined:
\begin{align}
s & \models true \nonumber \quad,\quad
s \models ap \quad\text{iff}\quad ap \in L(s)  \nonumber
\\
s & \models \neg \Phi \quad\text{iff}\quad s \not\models \Phi  \nonumber
\\
s & \models \Phi_1\wedge\Phi_2 \quad\text{iff}\quad s \models \Phi_1 \text{ and } s \models \Phi_2  \nonumber
\\
s & \models \mathcal{P}_{\bowtie p}(\Psi) \quad\text{iff}\quad Pr(s\models \Psi)\bowtie p \nonumber
\end{align}
$Pr(s\models \Psi)\bowtie p $ is the probability of the set of paths starting in $s$ and satisfying $\Psi$. Given a path $\psi$, if denote its \textit{i}-th state as $\psi[i]$ and $\psi[0]$ is the initial state. Then the satisfaction relation for a path formula for a path $\psi$ is defined as:
\begin{align}
\psi & \models X \Phi \quad\text{iff}\quad \psi[1] \models \Phi \nonumber
\\
\psi & \models \Phi_1 U^{\leq t}\Phi_2 \quad\text{iff}\quad \exists 0 \leq j \leq t \nonumber
\\
& (\psi[j]\models \Phi_2\wedge(\forall 0\leq k<j \; \psi[k]\models \Phi_1)) \nonumber
\end{align}


It is worth mentioning that both DTMC and PCTL can be augmented with rewards/costs, cf. \cite{filieri_probabilistic_2013}, which can be used to model, e.g. the energy/time consumption of the robot in a mission. Our approach is also compatible for those properties which we omit in this paper. 

By formalising the robot and its required properties in DTMC and PCTL respectively, the verification focus shifts to the checking of \textit{reachability} in a DTMC. In other words, PCTL expresses the constraints that must be satisfied concerning the probability, starting from the initial state, of reaching some states being labelled as, e.g. unsafe, failure or success. 
We use the tool PRISM \cite{kwiatkowska_prism_2011} which employs symbolic model checking algorithm to calculate the actual probability that a path formulae is satisfied (by extending the PCTL definition with $\mathcal{P}_{=?}(\Psi)$), then comparing with a required bound if given. 

\subsection{Parametric model checking}
PMC based on DTMC normally assumes the transition probabilities are known as constants which can be estimated from existing data and experts at design-time, or through system runtime monitoring. However, this traditional technique may not be suitable for runtime analysis in terms of the excessive time and power consumption, especially for robots. The idea of \textit{parametric} model checking (ParaMC), proposed by \cite{daws_symbolic_2005}, provides an efficient solution \cite{filieri_run-time_2011}. As shown by \cite{jansen_accelerating_2014}, modern tools can solve parametric DTMCs with thousands of states and transitions using reasonable computing resources, beyond the required capability of our methods here.

ParaMC can analyse DTMC whose transition probabilities are specified as functions over a set of parameters, e.g. Equation \eqref{eq_mdp_to_dtmc}. The result is then given as a closed-form rational function of the parameters which brings two advantages: (i) The verification can be divided into two steps. The computationally expensive symbolic analysis can be done at pre-mission stage when time and power constraints are not strong; then, during the mission, only substitutions are needed to replace the parameters in the closed-form expression with actual values learnt at runtime; (ii) Monotonicity analysis of the parameters can be easily done via the closed-form expression. Since, instead of point estimates, our new Bayesian estimator provides bounds for the parameters, thus monotonicity analysis is necessary to obtain meaningful bounds for the verification results.


\section{Estimates for DTMC transition parameters}
\label{sec_new_Bayesian_estimators}

\subsection{A fundamental Bayesian estimator}
\label{sec_fundamental_baye_est}

In a DTMC, given a current state $i$, the transition to a next state follows a \textit{categorical distribution}. Due to the Markov property, i.e. the choice of a next state only depends on the current one, the categorical distributions for each state are \textit{independent}. Hence, as we observe repeated transitions from the state $i$, the repeated categorical process follows a \textit{multinomial distribution}. Now the problem is reduced to the \textit{localised} learning of $k$ independent multinomial distribution, where $k$ is the number of states in the DTMC. From a Bayesian inference perspective, the posterior estimation requires a statistical model (the likelihood function) and a prior distribution. Note a complete description of Bayesian inference is beyond the scope of this paper.

For the $i$-th row of $\textbf{P}$, if we observe the data of transition number from state $i$ to $j$ as $n_{ij}$, and $n_i=\sum_{j=1}^{k}n_{ij}$ is the total number of outgoing transitions from state $i$, then the likelihood function is (by omitting the combinatorial factor which will be cancelled in the Bayesian formula):

\begin{equation}
Pr(\textmd{data} \mid p_{i1},...,p_{ik})=\prod_{j=1}^{k} p_{ij}^{n_{ij}}
\end{equation}

The method in \cite{epifani_model_2009} uses a Dirichlet distribution as priors for a given $i$-th row of $\textbf{P}$:
\begin{equation}
(p_{i1},...,p_{ik}) \sim Dir(n^{(0)}_i p_{i1}^{(0)},...,n^{(0)}_i p_{ik}^{(0)})
\label{eq_dirichlet_prior}
\end{equation}

where $n^{(0)}_i p_{i1}^{(0)},...,n^{(0)}_i p_{ik}^{(0)}$ are the \textit{canonical parameters} 
of the Dirichlet. $p_{ij}^{(0)}$ is the prior expectation for the transition probability $p_{ij}$, and larger $n_i^{(0)}$ leads to more concentrated probability measure around $p_{ij}^{(0)}$.
Thus, $n_i^{(0)}$ are quantifying the strength of beliefs in the prior $p_{ij}^{(0)}$, or a ``pseudo-count'' which can be interpreted as the size of an imaginary sample that gives the prior estimation \cite{walter_imprecision_2009}.

Then applying the Bayes rule, and thanks to both the conjugacy (with the multinomial likelihood function) and the canonical form, the posterior with updated parameters are:
\begin{align}
n^{(n_i)}_i&=n_i^{(0)}+n_i
\\
p_{ij}^{(n_i)}&=\frac{n_i^{(0)}}{n_i^{(0)}+n_i}\cdot p_{ij}^{(0)}+\frac{n_i}{n_i^{(0)}+n_i}\cdot\frac{n_{ij}}{n_i}
\label{eq_post_p_ij_fundamental}
\end{align}

Note, the upper index $^{(0)}$ is used to identify the prior parameters, in contrast, the $^{(n_i)}$ denotes the posterior parameters after observing $n_i$ outgoing transitions from the state $i$. As Equation \eqref{eq_post_p_ij_fundamental} shows, after seeing $n_{ij}$ out of $n_i$ as data, the posterior $p_{ij}^{(n_i)}$ is a \textit{weighted} sum of two terms: the prior estimate of $p_{ij}^{(0)}$ and the ${n_{ij}}/{n_i}$ which is the frequency of the relevant transition in the data. The weights are proportional to the $n_i^{(0)}$ (the ``pseudo-count'' of prior simple size) and $n_i$ (the ``actual-count'' of data sample size). Smaller $n_i^{(0)}$ represents lower confidence in the priors and the runtime data will dominate the posteriors. When $n_i^{(0)}\simeq 0$, \eqref{eq_post_p_ij_fundamental} reduces to the Maximum Likelihood Estimation \cite{epifani_model_2009}.

\subsection{A Bayesian estimator using sets of priors}
\label{sec_new_bay_sets_of_priors}
There are at least two practical issues with the method in Section \ref{sec_fundamental_baye_est}: (i) how to justify a particular choice of prior model parameters, i.e. $n_i^{(0)}$ and $p_{ij}^{(0)}$. In other words, whether one can truly express the subjective and imprecise prior knowledge with the \textit{exactness} that a particular prior distribution requires; and (ii) how to measure the error of single point estimation which will be propagated and compounded in the later model checking in ways that are unknown but likely to be significant, as pointed in \cite{calinescu_2016_formal}.

To address these two concerns, we utilise an imprecise probability 
model using sets of priors \cite{walter_imprecision_2009,walter_bayesian_2017} to model more vague prior knowledge by eliciting bounds of the prior parameters, and also resulting bounds for the posteriors whose range measures the estimation errors. Also importantly, it is sensitive to detect the \textit{prior-data conflict}, i.e. conflicts between prior assumptions and observed data \cite{evans_checking_2006}, which is useful to alter dangerous situations and will be discussed later.

To be exact, instead of a single value, we elicit an interval
for each prior parameters in \eqref{eq_dirichlet_prior} and denote these as:
\begin{equation}
n_i^{(0)} \in \left[\underline{n_i}^{(0)},\overline{n_i}^{(0)}\right]\quad , \quad p_{ij}^{(0)} \in \left[\underline{p_{ij}}^{(0)},\overline{p_{ij}}^{(0)}\right] \nonumber
\end{equation}

Then as proved in \cite{walter_imprecision_2009}, the lower and upper bounds for the posteriors of interest $p_{ij}^{(n_i)}$ are:
\begin{align}
\underline{p_{ij}}^{(n_i)}=\begin{cases} \frac{\overline{n_i}^{(0)}\underline{p_{ij}}^{(0)}+n_{ij}}{\overline{n_i}^{(0)}+n_i} & \text{if} \quad \frac{n_{ij}}{n_i} \geq \underline{p_{ij}}^{(0)}  \\ \frac{\underline{n_i}^{(0)}\underline{p_{ij}}^{(0)}+n_{ij}}{\underline{n_i}^{(0)}+n_i} & \text{if} \quad \frac{n_{ij}}{n_i} <\underline{p_{ij}}^{(0)}\end{cases}
\label{eq_post_lower_bound_pij}
\\
\overline{p_{ij}}^{(n_i)}=\begin{cases} \frac{\overline{n_i}^{(0)}\overline{p_{ij}}^{(0)}+n_{ij}}{\overline{n_i}^{(0)}+n_i} & \text{if} \quad \frac{n_{ij}}{n_i} \leq \overline{p_{ij}}^{(0)}  \\
\frac{\underline{n_i}^{(0)}\overline{p_{ij}}^{(0)}+n_{ij}}{\underline{n_i}^{(0)}+n_i} & \text{if} \quad \frac{n_{ij}}{n_i} >\overline{p_{ij}}^{(0)}\end{cases}
\label{eq_post_upper_bound_pij}
\end{align}

When $\frac{n_{ij}}{n_i} \notin \left[ \underline{p_{ij}}^{(0)},\overline{p_{ij}}^{(0)}\right]$, i.e. the prior-data conflict is at hand, the posterior interval $\left[\underline{p_{ij}}^{(n_i)},\overline{p_{ij}}^{(n_i)}\right]$ becomes wider, meaning we are \textit{even less certain} about the posterior estimation comparing to the priors. This is a new property comparing to other imprecise probability models in which the range of the posterior interval will \textit{always}, regardless of the prior-data conflict, decrease (i.e. converge to the observed data) as the sample size increases.

\subsection{Conservative Bayesian inference}

Catastrophic failures are modelled in our new framework by imposing transitions from unsafe states to catastrophic failures of different modes, e.g. the parameter $x$ in Fig.\ref{fig_DTMC} as an instance. The ``true unknown" values of these transition parameters may lie in very small orders of magnitude, say $10^{-5} \rightarrow 10^{-9}$, as a result of rigorous development process and safety-critical designs. This poses a great challenge for the estimation of such smaller failure rates in terms of that infeasible amount of statistical testing or operational time is required to observe sufficient failure data \cite{littlewood_validation_1993}. More practically, even if we did observe any catastrophic failures, we normally will redesign/update the robots, which makes the failure data obsolete. So effectively, we will only collect \textit{catastrophic-failure-free} data. Such ``good news'' would increase our confidence that the robot has a smaller chance to cause catastrophic failures, hence our claim is such inference has to be done in a \textit{conservative} way. The conservative Bayesian inference (CBI) method \cite{bishop_toward_2011,strigini_software_2013,zhao_conservative_2015,zhao_modeling_2017} was developed for safety-critical software to answer what can be claimed rigorously about the reliability when seeing failure-free runs. Here we introduce CBI as our Bayesian estimator for the catastrophic failure related parameters. Note, we only present the more practical case of seeing no catastrophic failures in this paper. The essential CBI can be extended to model very scarce failures.

In a given unsafe state $i$, as discussed above, the outgoing transitions follow a multinomial distribution, so \textit{marginally}, the number of transitions to the catastrophic failure state, say $n_{ij}$, is a binomial one, $n_{ij} \sim Bin(n_i,x)$ where $x$ is the transition probability and $n_i$ is the total number of outgoing transitions from state $i$. Then the likelihood function for the data of no catastrophic failure observed, i.e. $n_{ij}=0$, is: $Pr(\text{data}|x)=(1-x)^{n_i}$. So if we have a prior distribution for $x$, say $f(x)$, by Bayes rule we know the posterior expectation is:
\begin{equation}
E(x \mid \text{data})=\frac{\int_{0}^{1}x(1-x)^{n_i} f(x)\text{d}x}{\int_{0}^{1}(1-x)^{n_i} f(x)\text{d}x}
\label{eq_post_expected_x}
\end{equation}

Conventionally, starting from Equation \eqref{eq_post_expected_x}, we assign a parametric family for $f(x)$, e.g. a conjugate Beta in this case, or like the Dirichlet-Multinomial case in Section \ref{sec_fundamental_baye_est}. However, the use of conjugacy is based on the assumption that the practical situations we are dealing with have large quantities of failure data, in which the dominant contribution to posterior belief via Bayes Theorem comes from the likelihood function, i.e. situations in which ``the data can speak for themselves". We do not have this luxury for safety-critical systems with very limited or no failure data, so any use of a particular parametric family for $f(x)$ is questionable. 

Instead of assuming a \textit{complete} prior distribution that follows a parametric family, the assessors are more likely to have (or be able to justify) some very limited \textit{partial} prior knowledge, e.g. two possible scenarios: (i) ``I am 80\% confident the robot will not have any catastrophic failure in this unsafe state'' is a confidence in its catastrophic-failure-freeness, i.e. $Pr(x=0)=0.8$. See \cite{littlewood_reasoning_2012,strigini_software_2013} for the arguments of such partial prior knowledge; (ii) ``I am 90\% confident the probability of seeing a catastrophic failure from this unsafe state is smaller than 0.001'' is a confidence bound on a given probability of seeing catastrophic failure, i.e. $Pr(x \leq 0.001)=0.9$. Such partial prior knowledge could be supported by, e.g. when evidence is presented showing the system is strictly developed against IEC61508 SIL3\footnote{http://en.wikipedia.org/wiki/Safety\_integrity\_level.}.

In the above scenarios, the elicited partial prior knowledge is far from a complete prior distribution. Rather, if treat the partial priors as \textit{constraints} on a distribution, then there must be an \textit{infinite set of prior distributions} satisfying the prior constraints. Note, this set of priors is different from the one in Section \ref{sec_new_bay_sets_of_priors} which still assumes a parametric family (i.e. Dirichlet). Now the problem reduces to find the \textit{most conservative} prior distribution (in the sense of giving a maximum posterior expected transition probability) in the infinite set of priors satisfying the elicited prior constraints.

For example, as the first scenario, the assessor only has a $\theta$ confidence in its catastrophic-failure-freeness:
\begin{equation}
Pr(x=0)=\theta
\label{eq_cbi_perfection_prior_constraint}
\end{equation}
As proved in \cite{strigini_software_2013}, to maximise \eqref{eq_post_expected_x}, the corresponding $f(x)$, that subjects to the constraint \eqref{eq_cbi_perfection_prior_constraint}, is a two-point one with probability mass at $Pr(x=0)=\theta$ and $Pr(x=q)=1-\theta$, where $q$ is an optimisation point that can be obtained numerically. Thus, by such two-point $f(x)$, Equation \eqref{eq_post_expected_x} can be further bounded by:
\begin{align}
E(x \mid \text{data}) & \leq \frac{(1-\theta)q(1-q)^{n_i}}{\theta+(1-\theta)(1-q)^{n_i}} \nonumber
< \frac{(1-\theta)q(1-q)^{n_i}}{\theta} 
\\
& \leq \frac{(1-\theta)}{\theta (n_i+1)} (1-\frac{1}{n_i+1})^{n_i}
\label{eq_bounded_post_expected_x}
\end{align}
Note the last step above is because that $q(1-q)^{n_i}$ reaches its maximum when $q=1/(n_i+1)$.

Depending on the particular form of partial prior knowledge elicited, the worst-case priors varies, so does the posteriors. For cases of other forms of partial prior knowledge, see \cite{bishop_toward_2011,zhao_conservative_2015,zhao_modeling_2017}.

To summarise, in this section, we present two advanced Bayesian estimators for the two types of transition parameters. 
Mixed use of incorrect estimators will lead to either ``too conservative to be useful results'' or ``continuous prior-data conflict'', which we omit here due to the page limit.

\section{The framework with an UUV example}
\label{sec_frame_example}


\subsection{The running example}

\begin{figure}[ht]
	\centering
	\includegraphics[width=1\linewidth]{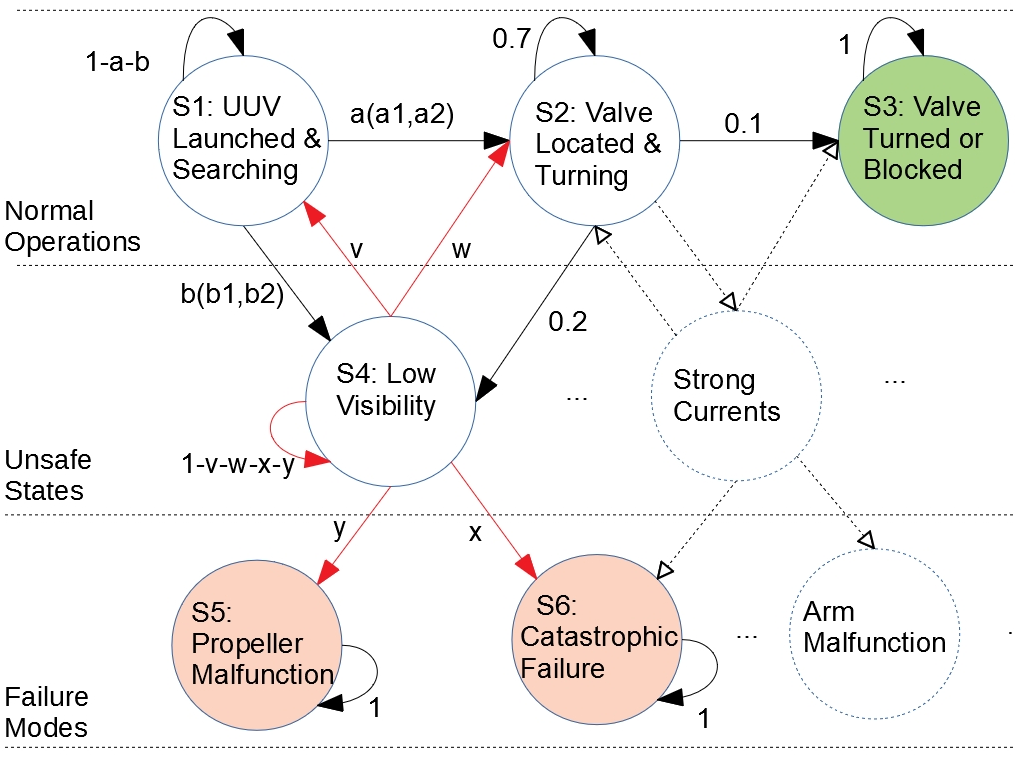}
	\caption{A layered DTMC modelling an UUV in a valve turning mission. The dotted shapes illustrate possible extensions to make the simplified example more realistic.}
	\label{fig_DTMC}
\end{figure}

We use an underwater search and valve turning mission in which a UUV equipped with electrical manipulators, stereo cameras, a specifically designed end-effector which had a camera in-hand with force and torque sensors to (i) locate a valve panel among different locations; and (ii) modify the valve handles to achieve different panel configurations. Possible disturbances are, e.g. muddy water, strong currents, blocked valves and panel occlusion. We formalise a single valve operation as an instance of the DTMC in Fig. \ref{fig_DTMC} which has been simplified with only 6 states, whilst more realistic formal models, e.g. with more states and transitions as shown as the dotted shapes in the figure, can be obtained in our proposed generic way as follows.

The DTMC in Fig. \ref{fig_DTMC} has 3 layers. The normal operation and safe states are modelled in the first layer, in which each state can transit into one or more unsafe states with some probabilities, e.g. in the $S1$ state of searching valve panel, the UUV may encounter a region of muddy water leading to the unsafe low visibility state $S4$. We group all the unsafe states in the middle layer which can either transit back to the normal operations or cause various modes of failures, including catastrophic ones, in the bottom layer, e.g. the propeller has a higher risk of malfunction when the UUV is in a muddy water. We believe the structure of a realistic DTMC can be built in a generic way that the top layer can be derived from the original policy-making model, e.g. MDP and reinforce learning as in \cite{PathakShashank2018Varo,pathak_ensuring_2013}, and the bottom two layers can be obtained from traditional safety and reliability analysis based on expert knowledge like Failure Mode, Mechanism and Effect Analysis (FMMEA) and Fault Tree Analysis (FTA). 

Practically, most of the transition probabilities in the formal Markov model can be argued as known constants, e.g. by prognostics on the remaining useful life of key components \cite{romain_review_2017}, leaving only a few transition probabilities as unknown parameters, e.g. the $x$ and $y$ in Fig. \ref{fig_DTMC} (thus a parametric DTMC), which will be Bayesian estimated at runtime with collected data. 

Fig. \ref{fig_DTMC} is also an \textit{induced} DTMC, since we assume the \textit{given} optimal policy is a probabilistic one for $S1$, e.g. with $\gamma$ probability to do speed-1 and $(1-\gamma)$ probability to do speed-2, where $\gamma$ is learnt by a separate primary planner. Then the induced transition probability from $S1$ to $S2$ is a function of $a_1, a_2$ as given by Equation \eqref{eq_mdp_to_dtmc}: $a(a_1,a_2)= \gamma a_1+ (1-\gamma) a_2$ where $a_1$ and $a_2$ are the probabilities of transiting from $S1$ to $S2$ with autonomous speed-1 and speed-2 respectively. Similarly for $b(b_1,b_2)$. For $S4$, we assume a deterministic policy (e.g. always do safe mode speed in low visibility), thus the outgoing transition probabilities are single parameters. 

We are interested in 4 requirements:
\textbf{R1:} What is the probability of completing a \textit{next} mission?  
\textbf{R2:} What is the probability of seeing catastrophic failures in a \textit{next} mission?  
\textbf{R3:} What is the probability of completing \textit{this} mission?  
\textbf{R4:} What is the probability of seeing catastrophic failures in \textit{this} mission? 
 The R1 and R2 will be analysed at the pre-mission stage, whose PCTL with the initial state specified as $S1$ are:
\begin{align}
P_{=?} \left[ \; true \; U (s=3) \right] \quad,\quad
P_{=?} \left[ \; true \; U (s=6) \right]
\nonumber
\end{align}
Whilst, R3 and R4 will be verified at runtime. Their PCTL are same as above but with different specified initial states.

\subsection{The framework and simulated experiments}

\begin{figure}[ht]
	\centering
	\includegraphics[width=1\linewidth]{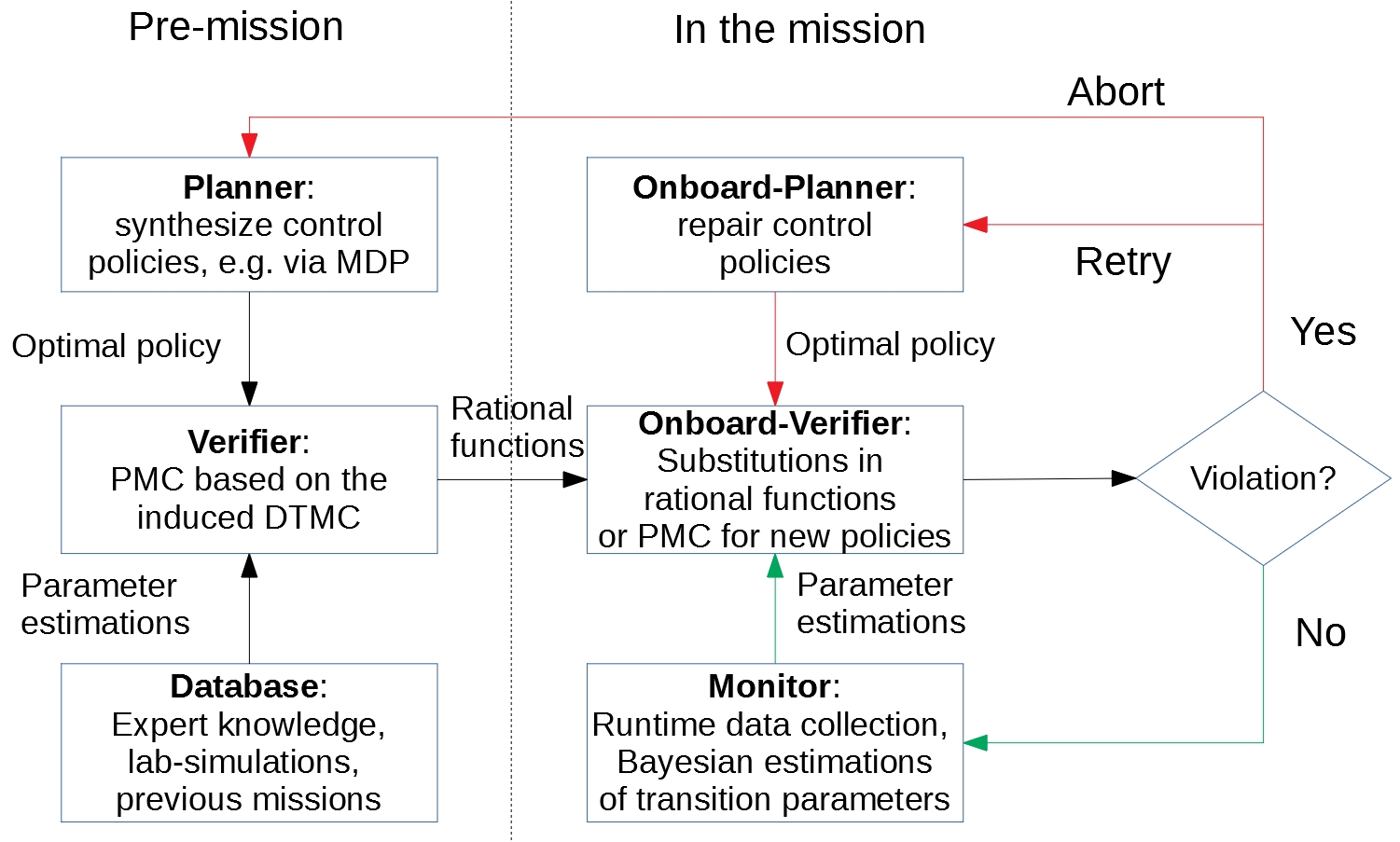}
	\caption{Overview of the verification framework.}
	\label{fig_framework}
\end{figure}

As shown in Fig. \ref{fig_framework}, our framework has two stages:

At the \emph{pre-mission stage}, a separate planner will first synthesise the optimal policy, which is assumed as given in the current framework. Given the optimal policy, an induced DTMC is derived and then revised to emphasise the safety and reliability by adding transitions to states representing hazards and various modes of failures, in which some transition probabilities are unknown parameters. Then ParaMC is conducted by tools like PRISM to generate rational functions for each PCTL property. The ParaMC results will be recorded and also reused by the on-board verifier to speed up the verification at runtime. Monotonicity analysis for each parameter in the rational functions is necessary due to the use of bounds for each transition parameter. Data analysis of experts' knowledge, lab-simulations and previous similar missions is performed to form prior knowledge required by the two Bayesian estimators. Finally, by substituting parameters in the rational functions with the prior estimates, pre-mission verification results are obtained, which represents our best efforts of assurance before launching the robot.

\emph{In mission (runtime)}, a monitor collects new data and does Bayesian estimates on the transition parameters. The on-board verifier checks if any required properties are being violated at runtime based on the ParaMC results at pre-mission stage. When any violation presents, actions like abort or instantly restart the mission (with a repaired policy) can be taken, whose implementation is beyond the scope of this paper. Otherwise, the UUV will continue and keep monitoring and verifying until the mission is complete.

We demonstrate and evaluate our framework by simulated experiments based on the UUV example of Fig. \ref{fig_DTMC}. First, to simulate and collect data of both previous missions and the current one for the pre-mission and runtime verifications respectively, we use the PRISM simulation module by assuming the \textit{true unknown} transitions probabilities in Fig. \ref{fig_DTMC} are: $a_1=0.05,a_2=0.03,b_1=0.2,b_2=0.1,x=10^{-5},y=0.001,v=0.3,w=0.01$. For the optimal policy of state 1, we assign a random number in $[\frac{25}{40},\frac{35}{40}]$ \footnote{All assumed values of the parameters in the simulation refer to the information provided by the PANDORA project.} for $\gamma$ in each simulated mission to mimic the practical case of changing policies. Then, it results in 49 previous missions and a 50-th as the current one\footnote{Without loss of generality, we choose a 50-th mission with relatively more transitions (522 to be exact) for a better illustration.} which explicitly using $\gamma=0.75$ as the optimal probabilistic policy of state 1. 


At the pre-mission stage, given the data of previous 49 missions (e.g. 1566 outgoing transitions from $S1$ with the action of speed-1, in which 1180 loops in $S1$, 85 to $S2$ and 301 to $S4$), together with the prior knowledge in Table \ref{tab_pre_mission_verification}, the posteriors by our Bayesian estimators are also listed there.  

\begin{table*}[ht]
\small
  \centering
  \caption{Prior knowledge and posterior estimations for the transition parameters given the data of 49 previous missions.}
    \begin{tabular}{|p{1.1cm}|c|c|c|c|c|c|c|c|}
      \hline   
      & $x$     & $y$     & $v$     & $w$     & $a_1$  & $b_1$  & $a_2$  & $b_2$ \\
    \hline
    pse. cou. & N/A   &  $[100,300]$     &    $[100,300]$   &   $[100,300]$    &   $[100,300]$    &   $[100,300]$    &    $[50,100]$   & $[50,100]$ \\
    \hline
    pri. est. &  $\theta=0.9$     & $[0.001,0.01]$      &   $[0.1,0.4]$    &    $[0.001,0.01] $  &   $[0.01,0.1] $   &  $[0.1,0.5]$   &    $  [0.01,0.1]$    &  $[0.1,0.5] $ \\
    \hline
    post. est. &  3.2e-5   &    $[0.0014,0.0032]$   &   $[0.27,0.33]$    &   $[0.0097,0.012]$    &   $[0.047,0.062]$    &    $[0.18,0.24]$   &    $[0.03,0.04]$   &  $[0.09,0.16]$\\
    \hline
    \end{tabular}%
  \label{tab_pre_mission_verification}%
\end{table*}%

Next, given $\gamma=0.75$ and using the ParaMC engine of PRISM, we obtain closed-form results for R1 and R2 (also R3 and R4 with different initial states for the use at runtime) as rational functions whose monotonicity can be easily analysed with respect to the 8 transition parameters. Then we obtain the pre-mission verification results:
\begin{equation}
R1 \in \left[ 0.857,0.961\right],\quad R2 \in \left[ 7.96\!\times\!10^{-4},1.55\!\times\!10^{-3}\right] \nonumber
\end{equation}

\begin{figure*}[ht]
\centering
\includegraphics[width=1\linewidth]{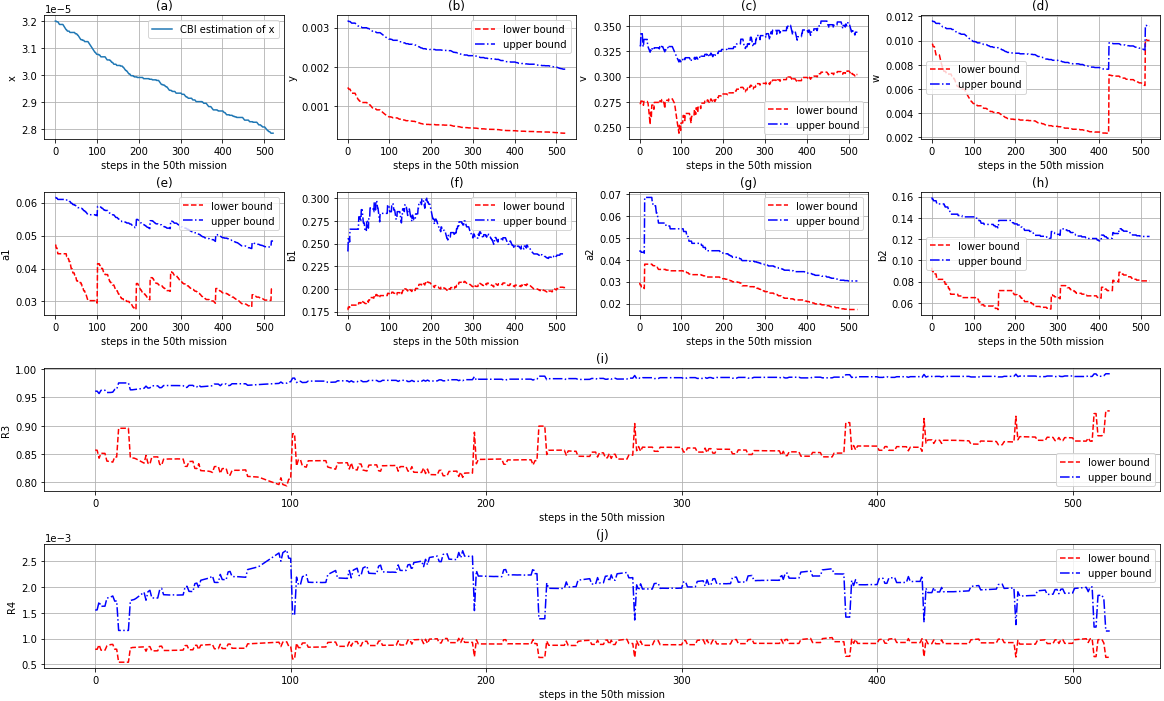}
\caption{Runtime Bayesian estimations for the 8 transition parameters (a)-(h) and verification results for R3 (i) and R4 (j).}
\label{fig_num_exam}
\end{figure*}

In the current 50-th mission, the posteriors of the 8 transition parameters in Table \ref{tab_pre_mission_verification} are in turn used as priors for the runtime Bayesian updates. The real-time estimates at each discrete time step of the transitions are plotted in (a)-(h) of Fig. \ref{fig_num_exam}, correspondingly R3 and R4 are plotted in (i) and (j).

As shown in the (a) of Fig. \ref{fig_num_exam}, the CBI estimation for $x$ decreases along with the execution of the robot without catastrophic failures. For non-catastrophic failure related parameters of (b)-(h), instead of using CBI to obtain a conservative point estimation, we plot both upper and lower bounds via Bayesian inference with sets of priors. Basically for each of these parameters, we first observe a divergence trend of the two bounds which then start to converge as more data is collected. Indeed, at the beginning of a mission, due to the sparse data collected, the prior-data conflict phenomenon is normal and foreseeable (e.g. 3 heads out of 3 tosses of a fair coin is conflicting our prior belief in its fairness). Such conflict is reflected by a wider bounded (i.e. less certain) estimation for each parameter and then propagated to the overall verification results. 

For instance, verifications on both R3 and R4 become less and less certain in the first 100 steps, partly due to the prior-data conflict of $a_1$. Then, the 101-th step is a transition from $S1$ to $S2$ with speed-1 action, which not only provides a new estimate of $a_1$ but also resolves the prior-data conflict to some extent, leading to tighter bounds for $a_1$, R3 and R4.

As more runtime data is observed, the general trend for each pair of bounds in Fig. \ref{fig_num_exam} starts to converge, meaning a more certain verification result. Note, both R3 and R4 highly depend on which state the UUV is currently in, so the obvious ``bumps'' in the plot for R3 are because the UUV is in $S2$ which certainly has a relatively higher chance to complete the mission. Similarly for the bumps in the plot of R4.

The above example is an ideal case in the sense that the new learning agrees with what it learnt before, i.e. all the 50 missions are simulated based on an MDP with the same transition probabilities, like the probability of transiting back to the normal state $S1$ from the low visibility state $S4$ is constantly assumed as $v=0.3$ for all 50 missions. For exceptional cases, e.g. there is a large body of muddy water implying $v=0.8$, our method will detect them as prior-data conflict through the whole mission, thus a continuous divergence of the bounds of related transition parameters will be observed, so consequently does an overall divergence of the verification results R3 and R4. We label this exceptional case as ``known unknowns'' in the sense that the new unknown (e.g. $v=0.8$) is learnt in an \textit{informed} way such that we know how much contradict it is to our prior knowledge, thus leading a less certain (i.e. wider bounds) new estimation.

We also would like to highlight the exceptional case of ``unknown unknowns'' in which the robot fails to know which state it is in and \textit{without knowing this fact}. For example, the UUV now is in a muddy water (i.e. $S4$), but the sensor fails to detect this unsafe state so that the UUV believes it is still in the normal state $S1$. As a consequence, the UUV does a wrong action of either speed-1 or speed-2 (i.e. the probabilistic policy in $S1$) in the unsafe state $S4$ instead of the safe speed (i.e. the deterministic policy for $S4$). Our method is also able to alert such dangerous case by detecting prior-data conflict phenomenon that happens \textit{simultaneously} for many parameters. Because, in the example above, although the UUV is \textit{actually} in $S4$, it will \textit{experience} constant loops in $S1$ without transiting to neither $S4$ (as the sensors fail to detect the abnormal states) nor $S2$ (as wrong actions are taken, assuming going fast in a muddy water will never detect the valves). Consequentially, prior-data conflict happens simultaneously for all $a_1,a_2,b_1,b_2$ parameters. 

More examples for the exceptional cases of ``known unknowns'' and ``unknown unknowns'' will be discussed in future due to the page limits here. We believe, detecting the prior-data conflict effect during the mission can assure protection against from faulty knowledge (i.e. epistemic limits) of the robot about its own state of health and environments.

\section{Related work}
\label{sec_related_work}

How should autonomous systems be verified is a new challenging question along with their increasing applications \cite{fisher_verifying_2013}. Formal methods must be integrated in order to develop, verify and provide certification evidence for large-scale and complex autonomous systems like robots \cite{farrell_robotics_2018}.

Model checking is a widely used formal method in verifying robotic systems, due to its relative simplicity and powerful automatic tools \cite{luckcuck2018formal}. For instance, in \cite{webster_generating_2014}, a proof-of-concept approach is presented to generate certification evidence for autonomous unmanned aircraft based on both model checking and simulation. PMC, as a variant, emphasises the inherent uncertainties of the formalised system. In \cite{konur_analysing_2012,gainer_probabilistic_2016}, the complex and uncertain behaviours of robot swarms are analysed by PMC. In \cite{norman_verification_2017,PathakShashank2018Varo}, PMC is used to verify the control policies of robots in partially unknown environments. In a hostile environment, the movements of adversaries are modelled probabilistically in \cite{cizelj_probabilistically_2011}. The reliability and performance of UUVs is guaranteed by PMC in \cite{gerasimou_efficient_2014}.

Although runtime PMC is effective for assuring the quality of service-based systems \cite{calinescu_dynamic_2011} and self-adaptive systems \cite{calinescu_2012_self-adaptive,filieri_probabilistic_2013}, there is little research on runtime PMC for robots. In the UUV domain, the first work of runtime PMC is credited to \cite{gerasimou_efficient_2014}. However, it focuses on improving the scalability of runtime PMC by using software engineering techniques, which is also applicable to our work here that focuses on developing new methods of learning model parameters.

One of the initial methods to learn the transition probabilities of DTMC is in \cite{epifani_model_2009}, which later has been retrofitted for CTMC \cite{filieri_formal_2012} and extended with ageing factors of collected data to accurately estimate time-varying transition probabilities \cite{calinescu_adaptive_2014}. To reduce the noise and provide smooth estimates, a lightweight adaptive filter is proposed in \cite{filieri_lightweight_2015}. Whilst, above mentioned approaches yield point estimations, these can be affected by unquantified and potentially significant errors. The work in \cite{calinescu_2016_formal} is the first to synthesise bounds for unknown transition parameters. However, it is based on the theory of simultaneous confidence intervals, which is fundamentally different to the Bayesian approach presented here which has the distinct advantage of being able to embed various forms of prior knowledge.

\section{Conclusions \& future work}
\label{sec_conclusion}

In this paper, we present a new framework to utilise PMC to assess the safety and reliability of robots at both pre-mission stage and runtime. 
Our main contributions are:
\begin{enumerate}
\item CBI is introduced with new closed-form results as a novel estimator for catastrophic failure related parameters.
\item Imprecise probability with sets of priors is introduced as another novel estimator for transition parameters. It allows to not only quantify the estimation errors and flexibly model imperfect prior knowledge, but also detect prior-data conflict to alter various dangerous situations.
\item A generic way is discussed to structure Markov models into layers to emphasise the system safety and reliability. 
\item A real-word application of UUVs has been formalised which can be extended and reused for future research.
\end{enumerate}

For illustration purpose, we demonstrate our methods with a stylised example which can be easily extended in our proposed generic way without changing the main conclusions, e.g. by considering the geographical waypoints in the top layer of the DTMC or listing complete failure modes in the bottom layer. The practicality of our new approach needs to be further evaluated with more case studies. We see potential issues like (i) insufficient safety analysis in FTA/FMMEA to generate sound DTMC in the bottom two layers; and (ii) unnecessarily complex DTMC model with too many unknown transition parameters which require too much prior knowledge from experts and too much data to be collected at runtime that burdens the on-board sensors.


We plan to exploit more use of prior-data conflict and also \textit{strong prior-data agreement} \cite{walter_sets_2016} to reflect the robot's epistemic limitations. Requirements to be verified for a robot should come from a higher level, e.g. when a robot is part of the Prognostics and Health Management (PHM) system, a PHM level PMC can be done in future to answer what to certify a robot. We also plan to propose a \textit{lightweight} on-board re-planner based on an MDP with the up-to-date bounded transition parameters.

\section{Acknowledgements}
This work was supported by the UK EPSRC, through the Offshore Robotics for Certification of Assets (ORCA) Hub [EP/R026173/1].	We thank Lorenzo Strigini, Peter Bishop and Andrey Povyakalo from City, University of London who provided insights on the initial ideas of the work.

\bibliographystyle{aaai}
\bibliography{ref}

\end{document}